%% file: main.tex
\documentclass{article} 
\usepackage{gpl_iclr2022_conference,times}

\input{math_commands.tex}

\usepackage{hyperref}
\usepackage{url}
\usepackage[T1]{fontenc}
\usepackage[final]{graphicx}
\usepackage{float}
\usepackage{subcaption}

\title{Reinforcement Learning for Location-Aware Scheduling}

\author{Stelios Stavroulakis\\
Department of Computer Science\\
University of California, Irvine\\
\texttt{sstavrou@uci.edu} \\
\And
Biswa Sengupta\\
Zebra Technologies\\
London, UK\\
\texttt{biswa.sengupta@zebra.com}
}

\iclrfinalcopy 
\begin{document}

\maketitle

\begin{abstract}
Recent techniques in dynamical scheduling and resource management have found applications in warehouse environments due to their ability to organize and prioritize tasks in a higher temporal resolution. The rise of deep reinforcement learning, as a learning paradigm, has enabled decentralized agent populations to discover complex coordination strategies. However, training multiple agents simultaneously introduce many obstacles in training as observation and action spaces become exponentially large. In our work, we experimentally quantify how various aspects of the warehouse environment (e.g., floor plan complexity, information about agents’ live location, level of task parallelizability) affect performance and execution priority. To achieve efficiency, we propose a compact representation of the state and action space for location-aware multi-agent systems, wherein each agent has knowledge of only self and task coordinates, hence only partial observability of the underlying Markov Decision Process. Finally, we show how agents trained in certain environments maintain performance in completely unseen settings and also correlate performance degradation with floor plan geometry.
\end{abstract}

\section{Introduction}

Recent successes in multi-agent reinforcement learning \citep{openai2019dota} have demonstrated super-human performance in imperfect information games with lengthy time-horizons, enormous state spaces, and sparse rewards. Scheduling problems with resource-dependent processing times are usually solved offline with linear programming relaxations \citep{linearprog} but are extremely slow in practice. As a learning paradigm, however, reinforcement learning is suitable for scheduling problems \citep{scheduling2019} and resource management \citep{deeprm} because, in principle, they are sequential decision-making problems with maximization of long-term reward as an objective. For that reason, recent online approaches using deep reinforcement learning \citep{deeprm} are implemented to trade-off optimality for computational efficiency and also dynamically adapt to unexpected events faster than a static schedule.

In modern warehouse environments, network outages might disrupt the communication links to/from robots. This incentivizes the use of decentralized multi-agent systems that can act independently and reliably. Each agent tries to self-schedule tasks to help reduce the average slowdown time of all tasks and contribute to the completion of tasks as efficiently as possible. Cooperation is required as sometimes the immediate reward might not be optimal for the team. 

This problem is intractable, in general. There have been multiple architectures proposed to capture this setting, such as the partially observable Markov Decision Process (POMDP) \citep{pomdp}. Usually, a POMDP in a $(n \times n)$ environment allows agents to only have access to a subset of the initial environment, usually in the form of a $(m \times m)$ grid where $m < n$. Sometimes, different states can translate to the same partial observation in a POMDP. This indistinguishability of states becomes evident in warehouse environments that are highly repetitive in nature, and an example of this is shown in Figure 1.

\paragraph{Our Contribution} In this paper we construct a different state and action space for warehouse scheduling applications in scenarios where task coordinates are available to the agents. This, in contrast with standard partial observability representations, allows rapid convergence with less computational overhead. We initially formulate the scheduling problem as a matching problem, exploit the location information available to the agents, consider a different state and action space and experiment with various existing algorithms. Furthermore, we also show that this lower-dimensional state and action space exhibits higher distinguishability between states, resiliency in floor plan perturbations, and generalizability in unseen environments.



\begin{figure}[htbp]
    \centering
    \includegraphics[width=0.25\textwidth]{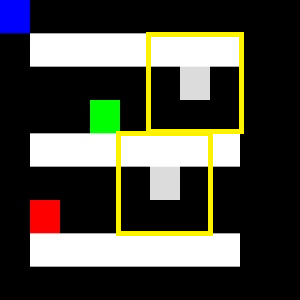}
    \caption{Gray agents with partial observability (yellow box) in a warehouse with isles. Agents are sensing the same input in multiple locations and are trying to navigate to the colorful targets.}
    \label{fig:my_label}
\end{figure}

\vspace{-0.4cm}
\section{Model}

In this section we describe a general model suitable to capture scheduling solutions in warehouse environments and lay down all implementation details of a custom OpenAI Gym environment \citep{openaigym} created for location-aware orchestration. In general, the environment contains agents and tasks that appear on a 2-dimensional map. Certain additive features were included to simulate environments closer to reality, including:

\begin{enumerate}
    \item Tasks are separated into two groups. One allows tasks to appear randomly on the map (non-stationary) and one constantly appears on the same location (stationary).
    \item Each task has a specific amount of work associated with its completion, translating to a specific amount of time required from the agents to be in the same coordinate and executing the task.
    \item Not all agents can execute all tasks. Each agent is instantiated with a certain action set.
    \item Some tasks are parallelizable and some are not. Figure 2 shows the capacity of each task, meaning that the acceleration of the completion of each task is clipped.
    \item A schedule is given in predetermined intervals, not at each time-step, meaning that no preemption is allowed. 
\end{enumerate}

\subsection{States}
The state-space can be seen in Figure 2. Preservation of as much information about the environment as possible while also minimizing the size of the observation was critical. If $N_A$ are the number of agents and $N_T$ the number of the available tasks, the complete state space is a $(N_A+1)\times N_T$ matrix. Each agent should not have access to the whole observation matrix and is fed only a fraction of this matrix, as communication is not allowed in decentralized settings. Every agent can only have access to its corresponding distance row and the work row, with a total observation space of $N_T \times 2$. Simply put, every agent can sense all distances to all tasks available, but not the distances of other agents to other target tasks. All distances and paths from agents to tasks are calculated using the A-star distance with the Manhattan distance heuristic \citep{astar}. 

\begin{figure}
\begin{subfigure}[b]{.49\textwidth}
  \centering
  \includegraphics[width=.9\linewidth]{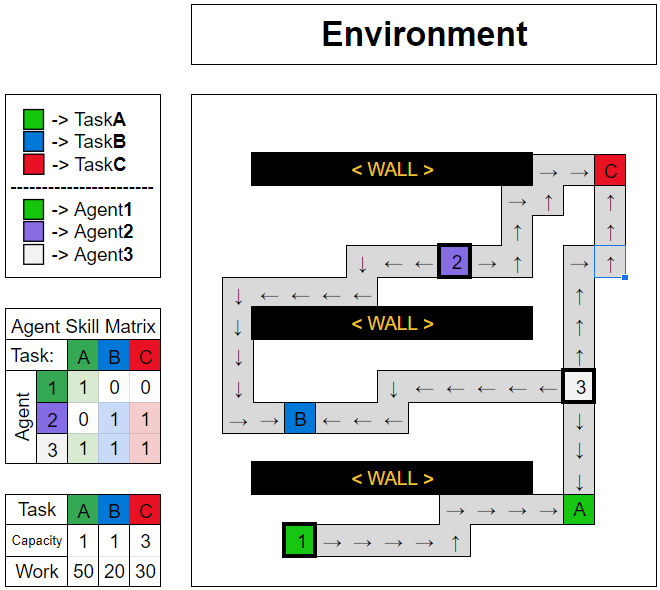}
  \caption{Environment representation. Agents and tasks have different colors, and each agent has a skill set (agent skill matrix). Before any action is taken, each agent considers the shortest paths to all tasks.}
  \label{fig:sfig1}
\end{subfigure}%
\hspace{2em}
\begin{subfigure}[b]{.49\textwidth}
  \centering
  \includegraphics[width=\linewidth]{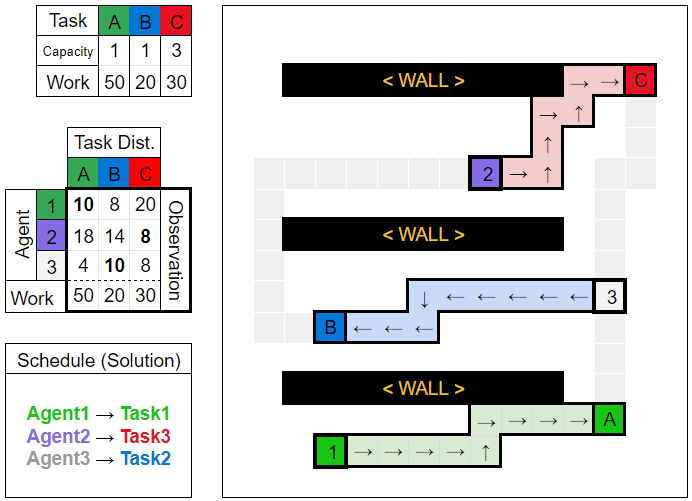}
  \caption{State space after pre-processing. Locations and skill sets are used to calculate distances from tasks. Agents are fed their corresponding task distance list as well as the work remaining for each task.}
  \label{fig:sfig2}
\end{subfigure}
\caption{Environment and state space representation.}
\label{fig:fig}
\end{figure}

\subsection{Actions}

When constructing an action space, it is necessary to ensure a rich and precise enough control to successfully solve a sequential decision-making problem. In this high-level scheduling approach, low-level planning is considered solved. We are, therefore, interested exclusively in selecting actions with dimensionality equal to the maximum number of tasks that could potentially appear in the environment $N_T$. Agents create a mapping of partially observable states to distributions over actions in order to select a task to navigate to. Agents should select an action based on their distance to the target and the amount of work remaining, potentially anticipating other agents completing tasks on their behalf and selecting tasks that make more sense for the sake of increasing reward.

When the environment is initialized, agents have certain inherent skills and can only execute tasks of a specific type. In Figure 2(a) we see how the Agent Skill Matrix determines which tasks can and can not be done by the agents. This means that effectively, agents have a modular action space. There have been numerous efforts creating neural network architectures that take care of modular action spaces \citep{pointernetworks}. However, in order to keep the architecture static, we only allow the agent to navigate to a task if the assigned task decision matches the corresponding task qualification of the agent. This is shown in Figure 3.

\begin{figure}[htbp]
    \centering
    \includegraphics[width=0.4\textwidth]{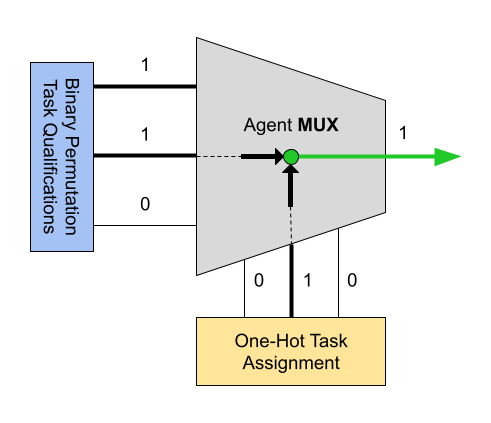}
    \caption{Successful example of task assignment. Task assignment matches the task qualification.}
    \label{fig:information_flow_world_algorithm}
\end{figure}

There are 3 necessary conditions in order for an agent to contribute to the completion of a task:\\ (1) Agent is assigned a task, (2) agent has skill to do the task and (3) task is not at max capacity.


\subsection{Rewards}

Reward shaping is the final component affecting environment design. Choices have to be made in terms of scale, magnitude, frequency (dense, sparse). Reinforcement learning agents strive for reward maximization, therefore, slight artifacts in the reward function might cause behavioral anomalies that resemble over-fitting in the strategy space rather than intuitive emerging behavior \citep{reward}. For scheduling applications, the average slowdown of all tasks is usually used as the main objective for optimization. Thankfully, there is an online version of the average slowdown time \citep{renato}. Our design allows for different priority tasks, resembling priority queues in the standard scheduling setup.

The reward signal of the environment needs to satisfy mainly two objectives

\begin{enumerate}
    \item Minimize average slowdown time of all tasks
    \item Prioritize task completion based on task priority
\end{enumerate}

Below is an example illustrating how different priorities affect action selection. In Figure 4 we see a set of 3 agents and 3 tasks, alongside the priorities of each task. When the priority of each task is the same, average slowdown time of each task is equally important. Hence, an optimal scheduler might assign \(Agent\_3\) to \(Task\_C\) in order to maximally utilize agents. However, if task A (for example) was a high-priority task, assigning \(Agent\_3\) to \(Task\_A\) would decrease the slowdown time of task A by the absolute value of the difference of the distances between \(Agent\_3\) to \(Task\_A\) and \(Agent\_1\) to \(Task\_A\), which would clearly increase the time it takes for \(Task\_C\) to be completed.

\begin{figure}[htbp]
\begin{subfigure}[b]{.49\textwidth}
  \centering
  \includegraphics[width=\textwidth]{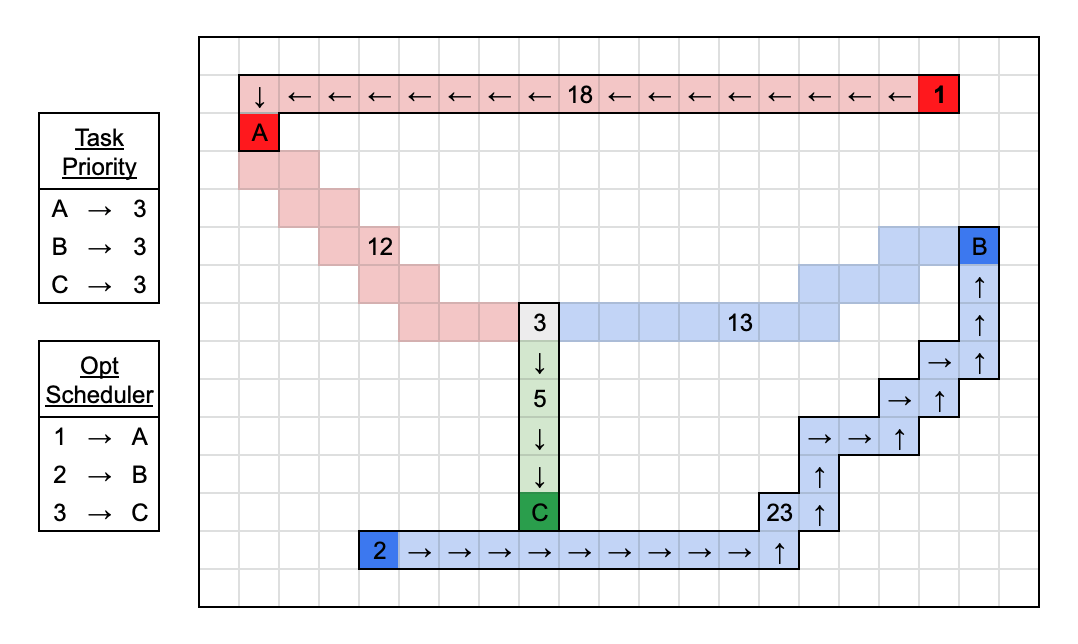}
  \label{fig:rew1}
\end{subfigure}%
\hspace{2em}
\begin{subfigure}[b]{.49\textwidth}
  \centering
  \includegraphics[width=\textwidth]{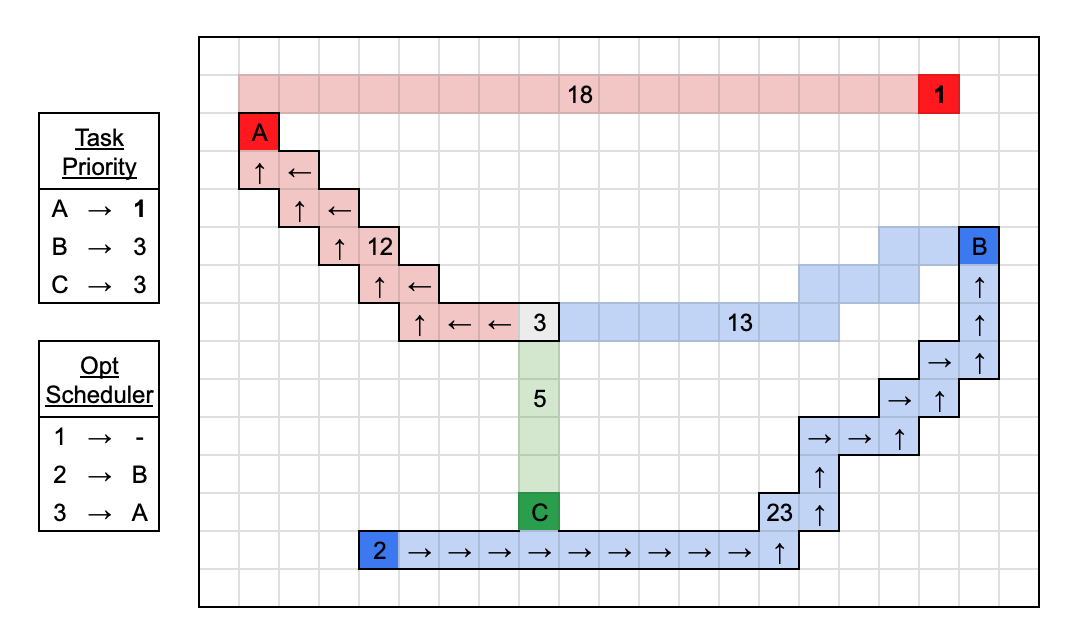}
  \label{fig:rew2}
\end{subfigure}
\caption{Different priorities yield different task assignments. In the left example. agent 1 is scheduled for task A. When A becomes high-priority in the right example, agent 3 is assigned to go to task A and abandon task C because it can reach it in less time-steps than agent 1.}
\label{fig:reward}
\end{figure}

\noindent The threshold in which \(Agent\_3\) is assigned \(Task\_A\) or \(Task\_C\) is mainly affected by the reward design. We introduce non-linearity in the reward function to break ties, since the reward is aggregated from multiple sources. Each task emits a negative reward at each time-step. Maximization of expected return, therefore, means completion of all tasks as fast as possible. It is also important to mention that a high-priority task should always be scheduled before a low-priority task, and the reward function has to encapsulate that information. For that reason, each task contains a tuple of variables \((r,p)\) that correspond to reward and priority. The total reward that the task emits to the environment at each time-step is:

\[ R(r,p) = -e^{-(p+1/r)} \]

Each task has a priority $p$ and a number $r$ associated with it, which scales the tasks' importance within the priority window. Due to the monotonicity of this function with respect to the terms $p$ and $1/r$, the reward function reassures a relative order with respect to priority and reward. For example, any task with priority $p=1$ will influence the sum of the reward more than any task with priority $p=2$, regardless of the value of $r$, since $\frac{1}{r} \in (0,1]$.

\section{Results}

\subsection{Max Flow}

\paragraph{Bipartite Matching} One naive approach of maximally utilizing agents would be to abstract this problem to a maximum bipartite matching problem \citep{karp} and solve it using max-flow \citep{maxflow}. Variations of this problem can handle assigning multiple agents to the same task with different capacities. We implemented this method to create a performance baseline for our results. Given a set of agents, tasks, and a compatibility matrix (connections of agents to tasks) one can assign agents to actions and maximally operationalize them. Below, we present the steps taken to model the orchestration problem as a max-flow problem.

\begin{figure}[htbp]
    \centering
    \includegraphics[width=250pt]{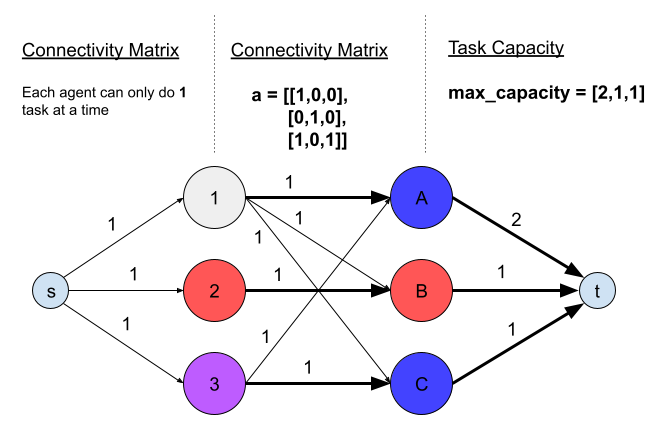}
    \caption{The Max-Flow Scheduler. This solution creates a schedule that uses the maximum number of available agents in the environment.}
    \label{fig:opt_scheduler_10}
\end{figure}

\paragraph{Max-Flow} The max-flow problem starts with assigning connections and weights from the source to the agents, the agents to the tasks, and the tasks to the sink. Source node should be connected to all agents and the weight is set to 1 because each agent can only do 1 task at the same time. This caps the flow from the source to the agent and allows for only one connection from the agent to any of the tasks that he is compatible doing. The connections from agents to tasks are determined by the compatibility matrix passed into the environment and weights are set to 1. Flow can not be larger than 1 due to constraint set from the source-agent connections. Finally, we connect all tasks to the sink node and assign weights equal to the maximum capacity of the tasks. This allows for multiple agents to be assigned a task, but no more than the capacity of the task. This solution guarantees maximal utilization of agents, meaning a maximum number of agents assigned to tasks. This solution is far from optimal because it does not consider the locations of agents and the time spent navigating to each task before completing it. Below is the average reward of different combinations of numbers of agents in an environment with three tasks and skills/agents.

\begin{table}[h!]
\centering
\begin{tabular}{||c c c c||} 
 \hline
 & 1 Agent & 2 Agents & 3 Agents \\ 
 \hline
 \hline
1 task/agent & -256.54 & - & - \\
2 tasks/agent & -180.23 & - & -  \\
3 tasks/agent & -123.86 & -74.11 & -40.60  \\ 
\hline
\end{tabular}
\caption{The table above shows the performance of the max flow solution in an environment where an increasing number of agents try to complete 3 different tasks in nature. In the first row each agent can only execute one out of the three tasks, in the second two out of 3 (in a circular fashion), and in the third row, agents can execute all three types of tasks. Max flow performance was evaluated by taking the average reward over 1000 episodes.}
\label{table:1}
\end{table}
\vspace{-0.4cm}
\subsection{Single-Agent PPO}
As a first step, we train a learnable agent using PPO \citep{ppo} in a T-shaped corridor environment. In the environment, we have an agent that can complete only one out of the three different tasks that might appear in this environment. The training parameters of the single-agent PPO learner are shown in Table 2. Figure 7 pictorially illustrates the setup.

\begin{table}[htbp]
\begin{center}
\begin{tabular}{ |p{100pt}|p{120pt}| }
\hline
\multicolumn{2}{|c|}{Training Parameters} \\
\hline
Task arrival probability & 0.5  \\
Workload & 10 \\
Agent1 Skills & Blue  \\
Agent1 Location & Random  \\
Reward    & Average task slowdown time \\
Value function clip & 15  \\
Total Iterations & 400k  \\
\hline
\end{tabular}
\end{center}
\caption{Training parameters of the single-agent PPO scheduler. Multiple copies of the same learn-able agent were deployed in the multi-agent case.}
\end{table}

\paragraph{Implementation Details} Tasks randomly appear in the corridor, and the agent can walk to any task in the corridor. Movement is phased. The agent is initially randomly placed in the corridor and at each time-step, a task appears with probability \(P(task) = c\). Decisions, however, are only made once every 10 time-steps. The agent has a modular set of available actions, as discussed in Section 2.2. It can either stay stationary or select one task from the available tasks and move towards it by leveraging A-star distance calculations. Masking out illegal actions were omitted due to the non-linearities it introduces to the problem. We approached the assignment of agents to non-existent tasks by heavily penalizing that action and keeping the agent stationary for that time-step. Once the agent selects a legal action, it does not necessarily mean that the agent is compatible with the task. Therefore, having the agent arrive at the task does not contribute to the completion of that task, hence does not affect the slowdown time of the task. Agents should develop the incentive to carefully select actions they are suitable with in order to minimize the average slowdown time of tasks. Figure 7 shows the training process of a single-agent PPO agent while in the corridor environment, decreasing the average slowdown time. The training curve also shows that the agent is consistently selecting the task it is able to execute since it is the only contribution it can make to lower the slowdown time. We visually verified convergence from various roll-outs at the end of the training.

\begin{figure}[htbp]
    \centering
    \includegraphics[width=200pt]{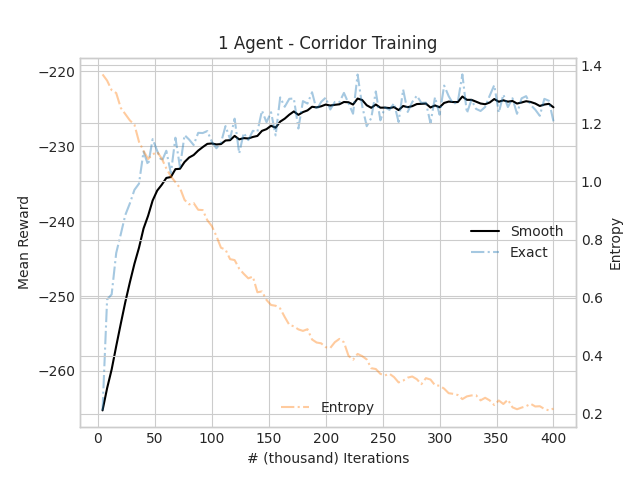}
    \caption{Mean reward and entropy during training in a T-shaped corridor, with one agent being able to execute only one out of three tasks randomly appearing in the environment.}
    \label{fig:opt_scheduler_1}
\end{figure}

\vspace{-0.4cm}
\paragraph{Outperforming Static Scheduler} We slowly increase the functionality of the agents and observe a consistently better performance. Being able to do more tasks means contributing to the decrease of the average slowdown time reduction. We also observe from Figure 8 that the increase in performance isn't linear. Orthogonal to increasing the capabilities of each agent, we fix the capability of the agent and bring more copies of the same agent into the environment in order to quantify how performance scales with respect to the controllable agents in the environment. Learning curves are shown in Figure 8. All skill/agent training curves at the end of training consistently beat the corresponding static maximum bipartite matching solution shown in Table 1.

\begin{figure}[htbp]
    \centering
    \includegraphics[width=0.15\textwidth]{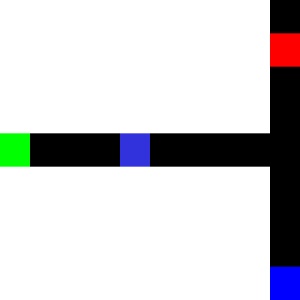}\hfill
    \includegraphics[width=0.15\textwidth]{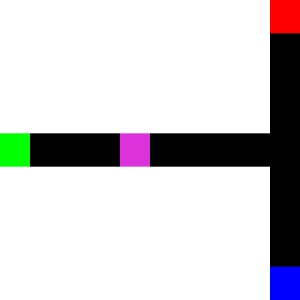}\hfill
    \includegraphics[width=0.15\textwidth]{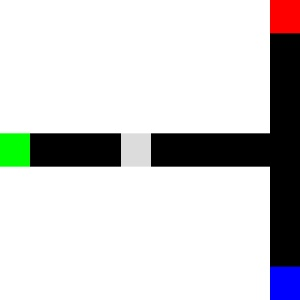}
    \caption{Corridor Environment. 3 different agents (blue/purple/gray) that can do a subset of (red/green/blue) tasks, based on their color combination.}
    \label{fig:opt_scheduler_2}
\end{figure}
\begin{figure}[htbp]
    \centering
    \includegraphics[width=0.49\textwidth]{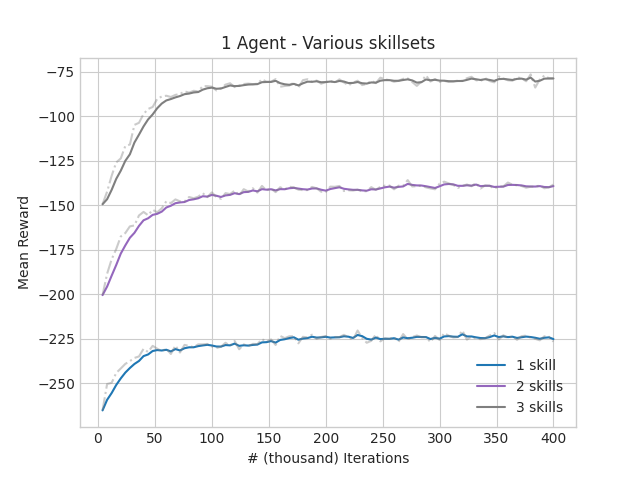}
    \includegraphics[width=0.49\textwidth]{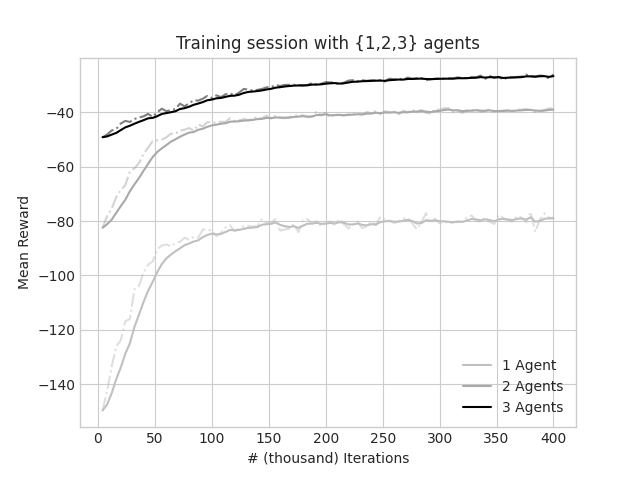}
    \caption{Graph of mean rewards during training. On the left, we showcase how average performance increases as our agent is allowed to execute more tasks. On the right, we deploy 1 or 2 or 3 identical agents in the same environment, all being able to execute all tasks.}
    \label{fig:opt_scheduler_3}
\end{figure}
\vspace{-0.1cm}
\subsection{Multi-Agent PPO}

Naturally, multiple identical agents taking actions simultaneously in an environment extends the framework to a Markov Game. The following results show the mean reward of multiple PPO trained agents after 400K time-steps considering 1,2 and 3 controllable units in the environment, independently trying to decrease the average slowdown time of all tasks, simultaneously. The rate in which the mean reward increases as the number of controllable units increases asymptotically approaches the theoretically maximum reward. We plot the average roll-out return of our trained policies for numerous controllable units. Table 10 demonstrates the non-linear increase in performance with respect to the total number of agents in the environment, which naturally occurs due to tasks reaching maximum parallelizability.

\begin{table}[h!]
\centering
\begin{tabular}{||c c c c c c||} 
 \hline
 & 1 Agent & 2 Agents & 3 Agents & 4 Agents & 5 Agents \\ 
 \hline
 \hline
Average Reward & -79.92 & -39.54 & -25.55 & -22.17 & -22.06 \\
\hline
\end{tabular}
\caption{Average reward w.r.t. number of agents in T-shaped corridor with 3 tasks and capacity of 5.}
\label{table:1124}
\end{table}
\vspace{-0.3cm}
\subsection{Transfer Learning Evaluation}


We test how agents trained in the corridor environment perform in the Grid-World environment. The performance is evaluated by comparing to the performance attained from agents trained from scratch in the Grid-World. Below, we present the more realistic 2D grid (resembling a warehouse) and plot the results of both the performance evaluation of a corridor trained model in this larger 2D grid environment and the training trajectory of a new model in the 2D environment.

\begin{figure}[h!]
    \centering
    \includegraphics[width=0.15\textwidth]{figures/1agent/3skill.png}
    \includegraphics[width=0.15\textwidth]{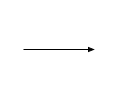}
    \includegraphics[width=0.15\textwidth]{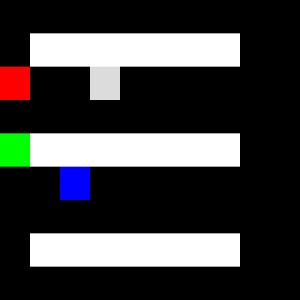}
    \caption{Agents trained in the Corridor environment are evaluated in the Grid-World environment.}
    \label{fig:opt_scheduler_222}
\end{figure}

\vspace{-0.1cm}
\paragraph{Resiliency} Results show that the pre-trained model in the simpler and smaller corridor environment almost matches the performance of the trained model in the new environment. This demonstrates resilience to perturbations in the floor plan, which can be explained by the representation of the observation space which contains minimal information about the floor plan structure, as well as the fact that both environments are for the most part convex. Knowledge is, therefore, transferable from one grid environment to another. Even though the transferability of the policy inevitably deteriorates as the complexity of the environment increases, the benefit that can be attained by training in smaller, simpler environments is worth the trade-off.

\begin{figure}[h!]
    \centering
    \includegraphics[width=0.15\textwidth]{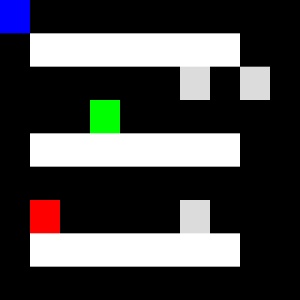}
    \includegraphics[width=0.15\textwidth]{figures/arrow.png}
    \includegraphics[width=0.15\textwidth]{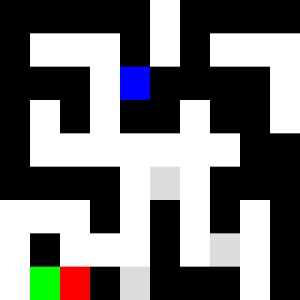}
    \caption{Agents trained in the Corridor environment are evaluated in the Grid-World environment.}
    \label{fig:opt_scheduler_212}
\end{figure}

\begin{figure}[h!]
    \centering
    \includegraphics[width=0.49\textwidth]{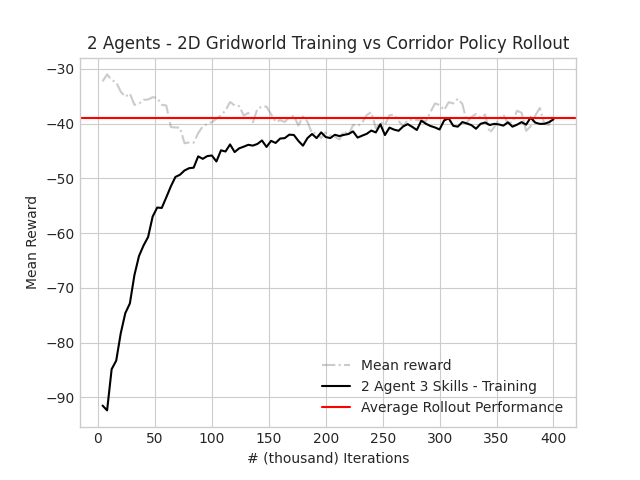}
    \includegraphics[width=0.49\textwidth]{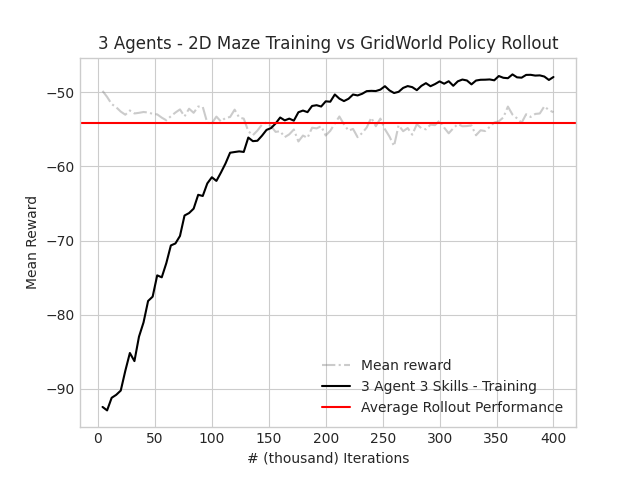}
    \caption{Left figure compares performance between the maximum roll-out performance of 2 agents trained in the corridor environment and training curve of PPO in Grid-World environment. Right figure shows how 3 agents trained in the Maze environment slightly outperform agents trained in the Grid-World environment and deployed in the Maze environment.}
    \label{fig:opt_scheduler_264}
\end{figure}
\vspace{-0.5cm}
\paragraph{Generalizability} When training in the Corridor environment, we know that the maximum $l_1$ distance between two points can only be $(width \times height)$. However, in the Maze environment, distances between agents and tasks increase. Hence, the generalization properties of neural networks become relevant when making intelligent decisions in an environment with larger and previously unknown combinations of distances.


\vspace{-0.3cm}
\section{Conclusion}

Modern warehouses are embracing decentralized operation of agents in their controlled environments.  We propose an efficient and compact representation of the spate and action space for the common setting of location-aware agents. This effectively converts a complicated task completion problem into a simpler scheduling problem, where the location of each agent affects the time delay before starting (and hence, also completing) a task. Distributing the average slowdown time as a common reward to all agents incentivizes co-operation and anticipation of future tasks based on their location. Furthermore, having the agents only select the high-level action to take (via the A-star algorithm), leads to extremely resilient performance to floor-plan perturbations, especially in grid-like environments, where the distance between points is equal to their Manhattan distance. Moving forward, we believe that our current formulation can be extended by using inverse RL instead of reward shaping. It would also be interesting to see how agents adapt to a task appearance distribution shift on the map. 

\bibliographystyle{gpl_iclr2022_conference}
\bibliography{gpl_iclr2022_conference}

\subsubsection*{Acknowledgments}
The author would finally like to thank Jay Williams and Bassam Arshad for their unparalleled support and Vaggos Chatziafratis for his constructive comments and feedback.

\end{document}

%% file: math_commands.tex
\usepackage{amsmath,amsfonts,bm}

\def\eqref#1{equation~\ref{#1}}

\def\1{\bm{1}}

\DeclareMathAlphabet{\mathsfit}{\encodingdefault}{\sfdefault}{m}{sl}
\SetMathAlphabet{\mathsfit}{bold}{\encodingdefault}{\sfdefault}{bx}{n}

